\documentclass[10pt,twocolumn,letterpaper]{article}
\usepackage{cvpr}              %

\usepackage{graphicx}
\usepackage{multirow}
\usepackage{array}
\usepackage{xcolor}  %
\usepackage{booktabs} %
\usepackage{makecell}
\usepackage{adjustbox}
\usepackage{amsmath}
\usepackage{caption}

\newcommand{\methodname}{MM-ACT\ }

\definecolor{cvprblue}{rgb}{0.21,0.49,0.74}
\usepackage[pagebackref,breaklinks,colorlinks,allcolors=cvprblue]{hyperref}

\title{MM-ACT: Learn from Multimodal Parallel Generation to Act}

\author{
Haotian Liang$^{1,4}$\thanks{Equal contribution.} \quad
Xinyi Chen$^{1,5}$\footnotemark[1] \quad
Bin Wang$^{1,6}$\footnotemark[1] \quad 
Mingkang Chen$^{3}$ \quad
Yitian Liu$^{2}$ \quad \\
Yuhao Zhang$^{2}$ \quad
Zanxin Chen$^{2}$ \quad
Tianshuo Yang$^{3,1}$ \quad
Yilun Chen$^{1}$ \quad
Jiangmiao Pang$^{1}$ \quad \\
Dong Liu$^{4}$ \quad
Xiaokang Yang$^{2}$ \quad
Yao Mu$^{2,1}$\thanks{Corresponding author.} \quad
Wenqi Shao$^{1}$\footnotemark[2] \quad
Ping Luo$^{3}$ \\[0.5em]
$^{1}$Shanghai AI Laboratory \quad
$^{2}$Shanghai Jiao Tong University \quad 
$^{3}$The University of Hong Kong \quad \\
$^{4}$University of Science and Technology of China\quad 
$^{5}$Fudan University \quad 
$^{6}$Zhejiang University \quad \\[0.25em]
}

\begin{document}
\setcounter{page}{1}
\pagestyle{plain}
\maketitle
\begin{abstract}

A generalist robotic policy needs both semantic understanding for task planning and the ability to interact with the environment through predictive capabilities. To tackle this, we present MM-ACT, a unified Vision-Language-Action (VLA) model that integrates text, image, and action in shared token space and performs generation across all three modalities. MM-ACT adopts a re-mask parallel decoding strategy for text and image generation, and employs a one-step parallel decoding strategy for action generation to improve efficiency. We introduce Context‑Shared Multimodal Learning, a unified training paradigm that supervises generation in all three modalities from a shared context, enhancing action generation through cross-modal learning.
Experiments were conducted on the LIBERO simulation and Franka real-robot setups as well as RoboTwin2.0 to assess in-domain and out-of-domain performances respectively. Our approach achieves a success rate of 96.3\% on LIBERO, 72.0\% across three tasks of real Franka, and 52.38\% across eight bimanual tasks of RoboTwin2.0 with an additional gain of 9.25\% from cross-modal learning. We release our codes, models and data at \url{https://github.com/HHYHRHY/MM-ACT}.

\end{abstract}

\thispagestyle{plain}  %
\pagestyle{plain}

\section{Introduction}

\label{sec:intro}

\begin{figure*}[th]
    \centering
    \includegraphics[width=\linewidth]{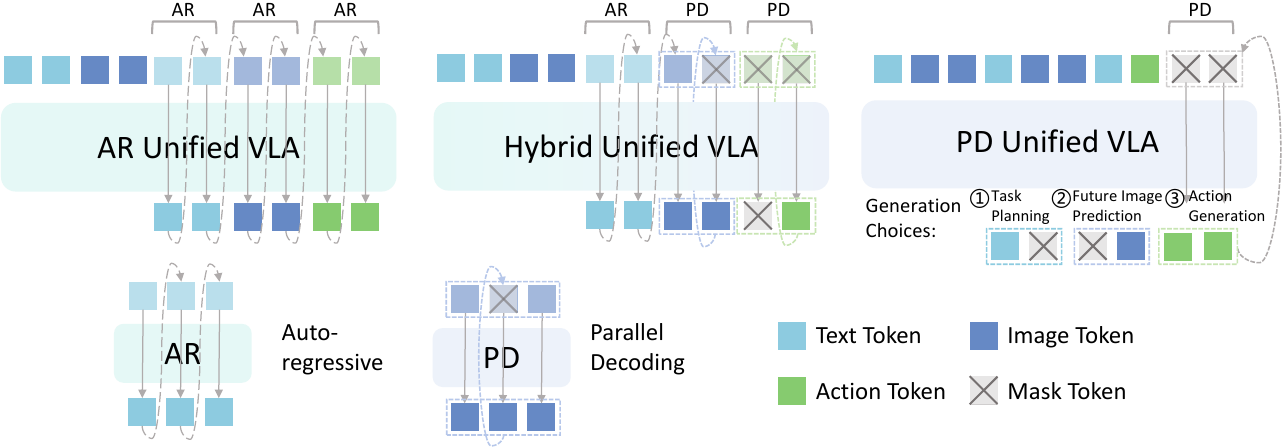}
    \caption{
        \textbf{Comparison of different unified VLA paradigms.} 
        (a) \textbf{Autoregressive (AR) unified VLA~\cite{zhang2025upvla}: }All three modalities—text, image, and action—are generated using an autoregressive approach.
        (b) \textbf{Hybrid unified VLA~\cite{wang2025univla, cen2025worldvla}: }Text is generated autoregressively, while image and action are generated using a re-mask parallel decoding approach.
        (c) \textbf{Parallel decoding (PD) unified VLA (Ours):} Converts images and instructions into a multimodal interleaved input, with text, image, and action generated through parallel decoding. This approach is capable of performing three tasks:\textcircled{1} sub-task planning, \textcircled{2} future image prediction, and \textcircled{3} action generation. Among them, \textcircled{1} \textcircled{2} adopt a multiple-step parallel decoding strategy with re-masking, while \textcircled{3} utilizes a one-step parallel decoding strategy.
    }
    \label{fig:motivation}
\end{figure*}

 A generalist robotic policy requires both high-level semantic understanding and the ability to interact effectively with the environment. In recent years, Vision-Language-Action (VLA) models~\cite{kim2024openvla, kim2025openvlaoft, black2024pi_0, pertsch2025fast, li2025cronusvla, lin2025onetwovla, yang2025instructvla} have emerged as a promising paradigm for building generalist robotic policies. These approaches typically build upon large-scale pretrained Vision-Language Models (VLMs) by integrating action heads or expert modules to bridge perception and control. However, while the underlying VLMs excel at visual and semantic understanding, they often lack an explicit model of physical dynamics~\cite{zhang2025vlabench, yang2025embodiedbench, gao2025genmanip}, which limits their ability to guide temporal action generation~\cite{zhao2025manipbench,chow2025physbench}.

Other works~\cite{hu2024vpp, tian2024seer, du2023unipi, zhu2024irasim, chen2025unified} further extend the conventional state-to-action imitation learning paradigm into a visual-prediction-driven decision and planning framework. By incorporating visual prediction into the policy learning process, these approaches enable models to explicitly or implicitly model future visual dynamics, thereby achieving stronger predictability and planning capabilities in complex interactive environments. Although these world models~\cite{blattmann2023svd, wan2025} excel in temporal and environmental dynamics, they are primarily trained for predictive objectives rather than task-oriented planning. This leads to limited instruction understanding and sub-task planning capabilities.%

Recent unified VLA approaches have largely inherited the development paradigm of unified understanding and generation models~\cite{deng2025bagel, team2024chameleon, xie2024showo,  wang2024emu3}. 
Instead of rethinking the policy architecture, these methods design action generation by closely following the base model’s modeling paradigm. For example, some works like~\cite{zhang2025upvla} retain the autoregressive text generation paradigm while adopting parallel decoding strategy for image and action generation, forcing the model to learn to handle single-token prediction and block-level tokens prediction in forward process. This in turn requires multiple attention mechanisms and substantially increases both architectural and training pipeline complexity. Others such as~\cite{cen2025worldvla, wang2025univla} adopt a fully autoregressive generation paradigm for text, image and action generation, which leads to slow inference speed for action generation.

To address these challenges, we propose MM-ACT, a unified model that jointly generates text, image, and action using a parallel decoding strategy. \methodname integrates text, image and action into a unified sequence of discrete tokens through modality-specific tokenizers, and learns to predict task planning, future image prediction and action chunk using block-level masked token prediction. For text and image generation, \methodname adopts a re-mask parallel decoding strategy, while for action generation it employs a one-step parallel decoding strategy to achieve low-latency inference. We further analyze the trade-offs between these two strategies in terms of effectiveness and efficiency. In contrast to prior approaches that rely on autoregressive text modeling, \methodname uses bidirectional attention over the full multimodal sequence, providing a more unified architecture that simplifies the training pipeline.

Building upon this unified modeling, we further propose \textbf{Context-Shared Multimodal Learning}, a training pipeline for three modalities generation. 
For identical multimodal context from robot's current views, task instructions, text descriptions and robot's states, we annotate data across the following categories: task planning, which involves planning the overall task and determining which sub-task to execute; future image prediction, consisting of image that represent the completion image of executed action chunk; and action chunk, referring to the chunk-step actions from expert data at the current timestep $t$.
After initial training steps on text and image generation, we further perform forward processes for all three modalities under the shared context, aggregating their losses within same gradient accumulation step for optimization.
During deployment, the model exclusively generates actions with one-step decoding strategy to enhance efficiency.

Our experiments aim to validate the feasibility of our modeling approach and the action generation's enhancement via our training pipeline. In simulation experiments, \methodname achieves an average of 96.3\% performance on Libero, 52.38\% in 8 tasks on RoboTwin2.0 and 72.0\% on Franka real-world experiments, consistently outperforming overall baselines. 
Moreover, our proposed training pipeline achieves a 9.25\% improvement in action generation performance compared to action-only learning baseline. Additionally, we demonstrate that either text-action or image-action joint training can also yield performance gains.

\section{Related Work}

\subsection{Discrete Diffusion Language Model}

Discrete diffusion models have recently emerged as promising alternatives to autoregressive language models. Unlike token-by-token decoding in AR models, diffusion-based approaches generate sequences through iterative denoising of corrupted inputs. Early models~\cite{austin2021structured, hoogeboom2021argmax} demonstrated the feasibility of applying discrete denoising to text. Subsequent work~\cite{sahoo2024simple} introduced improvements in masking strategies and training objectives. Large-scale bidirectional diffusion transformers~\cite{nie2025large, yang2025mmada} further advanced performance, achieving competitive results across text and multimodal tasks. These efforts establish discrete diffusion as a viable and increasingly competitive generation paradigm. In this work, we explore extending discrete diffusion models to the domain of action parallel decoding.

\subsection{Unified Vision-Language Models}
Autoregressive architectures have long been dominant in multimodal understanding, while diffusion-based models have become fundamental for image generation. Recently, there has been an increasing focus on developing unified frameworks~\cite{unifiedsurvey} that combine these tasks. Unified models' architectures can be abstracted as consisting of three components: modality-specific encoder, modality-fusion backbone, and modality-specific decoder. For autoregressive architectures~\cite{wang2024emu3, team2024chameleon, dong2023dreamllm, wu2025qwenimage, ge2024seedx}, both visual and language tokens are processed in a sequential manner, the modality-fusion backbone autoregressively predicts multimodal outputs. For diffusion-based architectures~\cite{yang2025mmada, wang2025fudoki, li2025dualdiffusion, shi2025muddit}, the denoising process is extended from timestep and noise to incorporate multimodal contexts, such as textual content, images, or joint embeddings. In hybrid autoregressive and diffusion architectures~\cite{zhou2024transfusion, xie2024showo, deng2025bagel}, text tokens are generated autoregressively, while image tokens are generated through a multi-step denoising process.  We aim to unify the training objectives and attention mechanisms for multimodal generation to simplify the training design. At the same time, we seek to ensure fast action generation by employing parallel action generation. 
Therefore, we have chosen MMaDA~\cite{yang2025mmada} as the base model.

\subsection{Vision-Language-Action Models}

The Vision-Language-Action (VLA) model primarily aims to transform general visual-language inputs into a sequence of executable action outputs within a unified framework. Early methods built upon large-scale pre-trained Vision-Language Models (VLMs) by incorporating action heads or specialized action expert modules~\cite{brohan2023rt1, zitkovich2023rt2, kim2024openvla, kim2025openvlaoft, black2024pi_0,pertsch2025fast, li2025cronusvla, song2025accelerating}. This allowed the model to simultaneously possess visual perception, language understanding, and action decision-making capabilities—achieving an end-to-end integration of vision, language, and action in one unified system. To bridge the latency gap between VLMs and the real-time demands of action systems, a dual-system design~\cite{lin2025onetwovla, bjorck2025gr00t, shi2025hirobot, zhai2025walloss, qu2025eo1} is commonly adopted, decoupling the overall architecture into an upstream VLM and a downstream action-prediction head. The latter either directly decodes actions from the latent action tokens generated by the VLM, or employs an additional diffusion-based action head to the VLM’s outputs or intermediate hidden representations. Various reasoning strategies~\cite{zawalski2024ecot, yang2025instructvla, chen2025internvlam1, deng2025graspvla, zhou2025chatvla} have also been used to augment the action model, ranging from purely language-based reasoning to multimodal reasoning capabilities. However, these approaches typically perform diffusion-based fine-tuning for action generation directly on top of autoregressive (AR) pre-trained backbones, resulting in an objective misalignment between AR-based pre-training (token prediction) and diffusion-based fine-tuning (denoising). This inconsistency can introduce optimization misalignment and hinder the model’s ability to effectively leverage its pre-trained knowledge.

In contrast to the above methods, our proposed model is built upon a diffusion LLM \cite{nie2025llada, yang2025mmada} (dLLM) as the foundation for the VLA. This design ensures that the model follows a consistent parallel decoding based generation objective during all training stages, thereby achieving better control and generalization capabilities.

\section{Method}
\begin{figure*}
    \centering
    \includegraphics[width=0.9\linewidth]{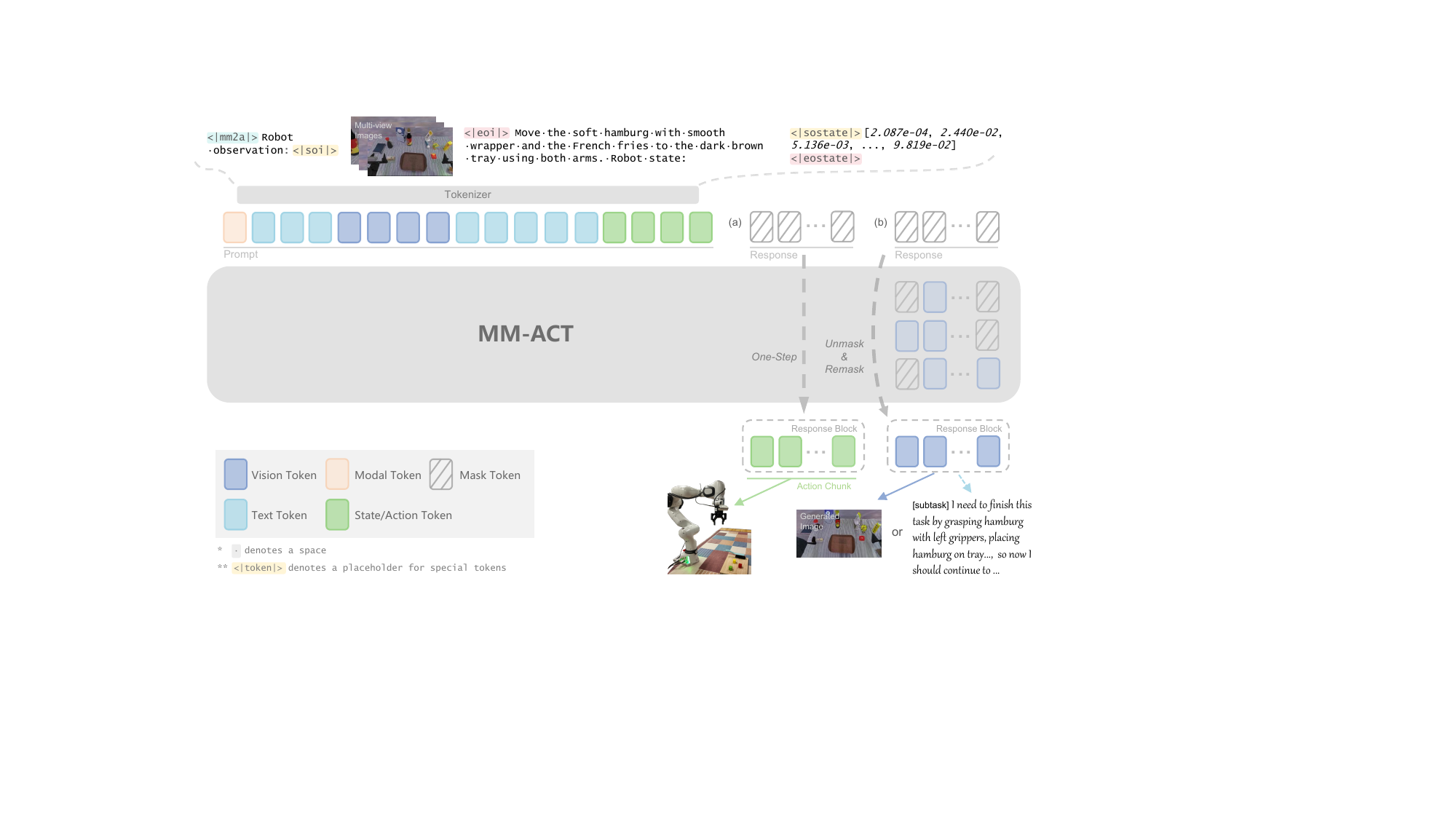}
    \caption{\textbf{The architecture of MM-ACT.} \methodname uses modality-specific tokenizers to tokenize text, image, and action into discrete tokens within a shared space. Given the shared multimodal input, the model determines whether to execute task planning, future image prediction, or action generation based on the modal token, with each task corresponding to the generation of text, image, or action.}
    \label{fig:method}
\end{figure*}

\subsection{Model Design}
We leverage a Transformer-based~\cite{vaswani2017attention} mask tokens predictor equipped with bidirectional attention mechanisms to facilitate generation tasks across three modalities.
Our model represents text, image and robot’s proprioceptive states as a single sequence of discrete tokens, drawn from the concatenated vocabularies of three modality-specific tokenizers. We add a modal token before the context that specifies the target generation modality, append a fixed-length $\texttt{<mask>}$ token block after the context. At inference time, we adopt different decoding strategies and compute probabilities at specific tokenizer positions corresponding to the modal token. Following MMaDA~\cite{yang2025mmada}, we use the LLaDA~\cite{nie2025llada} model’s tokenizer for text and employ pretrained image quantizer from Show-o~\cite{xie2024showo}, encoding and decoding image inputs and outputs, which uses 8,192 tokens as the image codebook. For image inputs, we first pad each image to a square, downsample it to 256×256 and encode it into 256 tokens. For image generation, the model outputs 256 tokens and the image quantizer decodes into a 256×256 image. For robot's states inputs and action outputs, we adopt the bin tokenizer~\cite{kim2024openvla} as the quantization method, allocating 2,048 tokens dedicated to action generation. On the input side, each continuous scalar is first normed to the range of [-1, 1], then is quantized to a token in the codebook; on the output side, tokens are detokenized back into continuous scalars to represent continuous action values.
We concat action codebook to the end of the tokenizers, without affecting the original text tokenizer and image codebook.

\subsection{Context‑Shared Multimodal Learning}

\paragraph{Context‑Shared Multimodal Input}
We use a shared context for the generation tasks across the three modalities. The context for each modal 
\( C_{\mathrm{modal}}\! =\! \text{\texttt{<modal>}}\! +\! \text{\texttt{sharedinput}}\), where the modal token 
\( \text{\texttt{<modal>}}\!\in\!\{\text{\texttt{<|mm2a|>}},\text{\texttt{<|mmu|>}},\text{\texttt{<|t2i|>}}\} \).
$\texttt{shared\-input}$ is a modality-interleaved token sequence following a template to convert the inputs of robot's multi-view observations, task instructions, text descriptions, and optionally the robot's states.

We append a fixed-length block for image and action modals after $C_{\mathrm{modal}}$. 
Specifically, the text block size is set to 256 to accommodate task planning sequence. We concatenate variable-length textual annotations after the context and append $\texttt{<eos>}$ tokens to match the maximum sequence length. The image block size is also 256, enabling the generation of a single image.
The action block size $N_{\mathrm{act\_block}}=d_{\mathrm{action}}*N_{\mathrm{chunk\_size}}$, where $d_{\mathrm{action}}$ is the dimension of action and $N_{\mathrm{chunk\_size}}$ is the number of actions in one chunk. It enables generation of one chunk of actions, with $N_{\mathrm{chunk\_size}}$ kept fixed during both training and inference.

We set the same maximum sequence length for the inputs of all three modalities, and shorter sequences are padded with the $\texttt{<pad>}$ token to match this maximum length.
\vspace{-1em}
\paragraph{Multimodal Learning with Unified Objective} 
The generation tasks across the three modalities share a common context, and we append modality-specific blocks with mask tokens after the context to train generation capability. Unlike approaches that combine multiple objectives, such as autoregressive text generation and diffusion-based generation for images~\cite{zhou2024transfusion} or actions~\cite{wen2025dexvla}, we adopt the same optimization objective to train generation across all three modalities.
We model the specific block of each modality as a token sequence $x_{0} = \bigl(x_{0}^{1},\,x_{0}^{2},\,\dots,\,x_{0}^{L}\bigr)$, $L$ is the sequence length. For a continuous time $t\in(0,1]$, construct a masked sequence $x_{t}$ by independently masking each position with probability $p_{\mathrm{mask}}=f_{\mathrm{modal}}(t)$, where $f_{\mathrm{modal}}$ is the function of mask schedule in each modal. The conditional distribution of $x_t^i$ can be formed as:
\begin{equation}
\begin{aligned}
q_t(x_t^i \mid f_{\mathrm{modal}}(t), x_0^i)
={}& (1 - f_{\mathrm{modal}}(t))\,\mathbf{1}\{x_t^i = x_0^i\} \\
&+ f_{\mathrm{modal}}(t)\,\mathbf{1}\{x_t^i = \texttt{<mask>}\}.
\end{aligned}
\end{equation}
We use linear schedule for text modality followed by LLaDA~\cite{nie2025llada} and cosine schedule for both image and action modalities to match the noise schedule for continuous denoising. Then the conditional distribution of $x_{t}$ can be formed as:
\begin{equation}
q_{t}(x_t \mid x_0) = \prod_{i=1}^{L} q_{t}(x_t^{i} \mid f_{\text{modal}}(t), x_0^{i})
\end{equation}
We set $t=1$ for action modal to train the model generating all tokens from a fully masked sequence in a single forward pass, the masked sequence $x_{t}=\texttt{<mask>}\times L$.

Formally, \methodname is trained as a mask token predictor, a parametric model $p_\theta(\cdot \mid C_{\mathrm{modal}}, x_t)$ that takes $C_{\mathrm{modal}}$ and $x_t$ as inputs and predicts all masked tokens simultaneously. We define a unified cross-entropy loss on masked tokens within three modal.  

\begin{figure*}[htbp]
    \centering
    \includegraphics[width=0.9\linewidth]{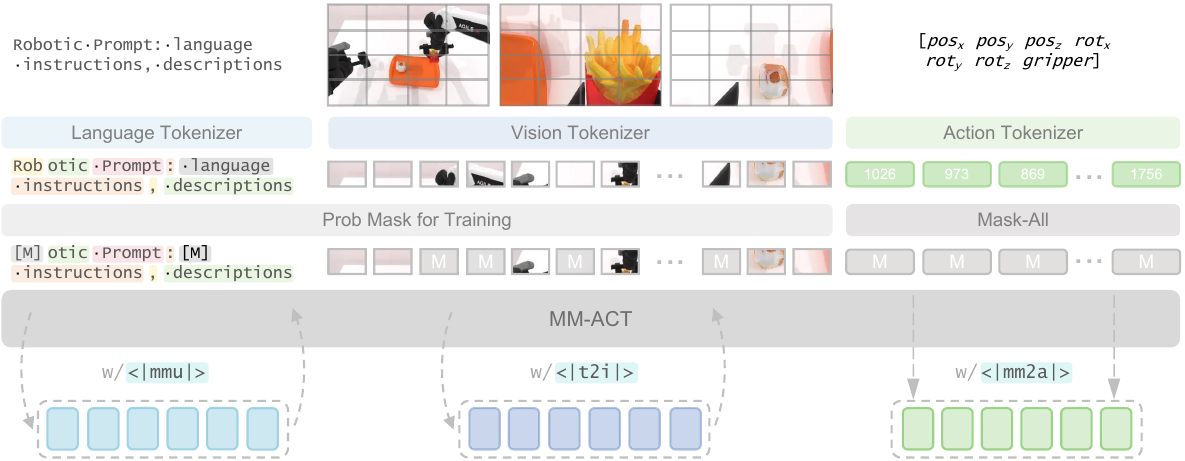}
    \caption{\textbf{Training pipeline of MM-ACT.} Within the shared context, ground truth of three modalities are masked according to the decoding strategy of each modality, and appended to the context. The model takes these inputs to perform forward processes across three modality generation tasks, computing the loss specifically on the masked tokens.}
    \label{fig:method-supplementary}
\end{figure*}
\begin{equation}
\label{eq:unified-loss}
\begin{aligned}
\mathcal{L}(\theta) ={}& -\mathbb{E}_{t,x_0,x_t}\Big[
\sum_{\mathrm{modal}\in\mathcal{M}} \frac{\lambda_\mathrm{modal}}{t}
\sum_{i\in\mathcal{I}_\mathrm{modal}} \mathbf{1}\{x_t^i=\mathrm{M}\} \\
&\qquad\qquad \times \log p_\theta(x_0^i \mid C_{\mathrm{modal}}, x_t)
\Big].
\end{aligned}
\end{equation}
where $x_0$ is sampled from the training data, $x_t$ is sampled from $q_{t}(x_t \mid x_0)$, $\mathrm{M}$ is $\texttt{<mask>}$ token. 
The indicator function $\mathbf{1}[\cdot]$ ensures that the cross-entropy loss is computed only for masked tokens. 
$\mathcal{M} = \{\texttt{<|mm2a|>}, \texttt{<|mmu|>}, \texttt{<|t2i|>}\}$, 
and $\lambda_\mathrm{modal}$ represents the weight of each modal's loss, used to control the impact of each modal on model optimization during training process. In Figure~\ref{fig:method-supplementary}, we provide an illustration of our mask data construction process during training.

We adopt a two-stage training strategy. In Stage 1, we set $\lambda_\mathrm{mm2a}$ to $0$ and initially train the model exclusively on text and image generation tasks, optimizing until the losses for text and image modalities reach low values. In Stage 2, we primarily focus on supervising action generation, adjusting $\lambda_\mathrm{mmu}$ and $\lambda_\mathrm{t2i}$ to approximately $0.05–0.1$ to maintain their generative capabilities.

\begin{table*}[htbp!]
\centering
\begin{adjustbox}{width=\linewidth} 
\begin{tabular}{>{\raggedright}m{5.5cm} c c c c  c}
\toprule

\textbf{Model} & \textbf{Spatial SR (\%)} & \textbf{Object SR (\%)} & \textbf{Goal SR (\%)} & \textbf{Long SR (\%)} & \textbf{Average SR (\%)} \\
\midrule
OpenVLA \cite{kim2024openvla} & 84.7 & 88.4 & 79.2 & 53.7 & 76.5 \\

$\pi_0$ + FAST \cite{pertsch2025fast} & 96.4 & 96.8 & 88.6 & 60.2 & 85.5 \\
$\pi_0$ \cite{black2024pi_0} & 96.8 & \underline{98.8} & \underline{95.8} & 85.2 & 94.2 \\
OpenVLA-OFT \cite{kim2025openvlaoft} & 96.2 & 98.3 & \textbf{96.2} & 90.7 & 95.4 \\
\midrule
CoT-VLA \cite{zhao2025cotvla} & 87.5 & 91.6 & 87.6 & 69.0 & 81.1\\
TraceVLA\cite{zheng2024tracevla} & 84.6 & 85.2 & 75.1 & 54.1 & 74.8 \\
DreamVLA \cite{zhang2025dreamvla} & \underline{97.5} & 94.0 & 89.5 & 89.5 & 92.6 \\
\midrule
WorldVLA (512*512) \cite{cen2025worldvla} & 87.6 & 96.2 & 83.4 & 60.0 & 81.8 \\
UniVLA \cite{wang2025univla} & 95.4 & \underline{98.8} & 93.6 & \textbf{94.0} & \underline{95.5} \\
\midrule
\textbf{\methodname (Vanilla) } & \textbf{97.8} & \textbf{99.4} & 94.8 & 88.0 & 95.0 \\
\textbf{\methodname (+Text in Long)} & - & - & - & \underline{93.0}(\textbf{\textcolor{blue}{+5.0\%}}) & \textbf{96.3} \\
\bottomrule
\end{tabular}

\end{adjustbox}
\caption{\textbf{LIBERO task performance results (\%).} \textbf{Bold} values denote the best performance, and \underline{underlined} values denote the second-best.}
\label{tab:libero}
\end{table*}

\paragraph{Parallel Decoding Strategy}
We formulate the model as a block-level masked-token predictor. For action generation, we produce all action tokens in a single forward pass to maintain efficiency. We also introduce re-mask decoding strategy for action generation in our design, adopt a low-confidence remasking strategy and use a cosine noise schedule consistent with MAGVIT-v2~\cite{magvitv2}. We compare effectiveness and efficiency in \ref{parag:action decoding strategy}. We employ re-mask decoding strategy for image generation, with same re-mask strategy and noise schedule to action.

For text generation, we limit the generation sequence length to 256 tokens. This constraint is adopted because task planning annotations in our manipulation tasks can typically be completed within this range. Furthermore, this limit aligns with the default block size of LLaDA~\cite{nie2025llada}, allowing us to restrict the entire generation process to a single block.
As a result, we do not employ the semi-autoregressive approach for text generation. Instead, text, image, and action generation are all performed within one block. Logits are computed for all masked positions, and a subset of tokens selected either randomly or based on confidence scores is predicted. The masking schedule is linear, and the forward process is repeated for a fixed number of steps. More details can be found in Appendix~\ref{app-sec:details of remask}.

\section{Experiments}

\begin{table*}[ht!]
\centering
\begin{adjustbox}{width=\linewidth} 
\begin{tabular}{lcccc}
\hline
{\textbf{Model}} & \textbf{Adjust Bottle}&\textbf{Beat Hammer Block} & \textbf{Click Bell} & \textbf{Dump Bin Bigbin}  \\
\hline
$\pi_0$ \cite{black2024pi_0} & \textbf{89\%} & \underline{68\%}  & 40\%  & \textbf{61\%}  \\
OpenVLA-OFT \cite{kim2025openvlaoft} & 64\%  & 7\%  & 24\%  & \underline{31\%} \\
\textbf{\methodname(Vanilla)} & 51\%  & 61\%  & 86\%  & 13\%   \\
\textbf{\methodname(+Text)} & \underline{75\%}  &  67\%  & \underline{91\%}  & 13\%  \\
\textbf{\methodname(+Image)} & 72\%  & 64\%  & \underline{91\%}  & 8\%  \\
\textbf{\methodname(+Text\&Image)} & 71\%  & \textbf{78\%}  & \textbf{95\%}  & 13\%  \\
\hline
{\textbf{Model}} & \textbf{Move Playingcard Away}&\textbf{Lift Pot} & \textbf{Place Burger Fries} & \textbf{Place Can Basket} \\
\hline
$\pi_0$ \cite{black2024pi_0}            & \textbf{47\%}  & 29\%  & 41\%  & 10\%  \\
OpenVLA-OFT \cite{kim2025openvlaoft}    & 17\%  &  3\%  & 31\%  & 8\%  \\
\textbf{\methodname(Vanilla)}              & 30\%  & \textbf{40\%}  & 46\%  &  \underline{18\%}  \\
\textbf{\methodname(+Text)}   & 24\%  &  28\%  & 56\%  & \underline{18\%} \\
\textbf{\methodname(+Image)}  & \underline{39\%}  &  \underline{31\%}  & \underline{72\%}  & 13\%  \\
\textbf{\methodname(+Text\&Image)}  & \underline{39\%}  &  \underline{31\%}  & \textbf{73\%}  & \textbf{19\%}  \\
\hline
\multicolumn{5}{c}{\textbf{Overall Avg:}\hspace{10mm} $\pi_0$ = 48.13\%, \hspace{10mm} OpenVLA-OFT = 23.13\%, \hspace{10mm} \methodname(Vanilla) = 43.13\%} \\
\multicolumn{5}{c}{\methodname(+Text) = 46.5\%(\textbf{\textcolor{blue}{+3.37\%}})\hspace{10mm} \methodname(+Image) = \underline{48.75\%}(\textbf{\textcolor{blue}{+5.62\%}})}\hspace{10mm} \methodname(+Text\&Image) = \textbf{52.38\%}(\textbf{\textcolor{blue}{+9.25\%}})\\
\hline
\end{tabular}
\end{adjustbox}
\caption{\textbf{RoboTwin task performance results.} \textbf{Bold} values denote the best performance, and \underline{underlined} values denote the second-best.}
\label{tab:robotwin}
\end{table*}

Our experiments center around two primary questions: (1) whether our model architecture can effectively perform action generation to accomplish manipulation tasks in both in-domain and out-of-domain settings; and (2) what kind of improvements our training pipeline provides across generation tasks in different modalities.

In \ref{sub:benchmark},we describe the benchmarks and experimental setups used in both simulation and real-robot environments. \ref{sub:implementation details} introduces the datasets and training details. \ref{sub:evaluation_results} presents the evaluation results for action generation. \ref{sub:training_analysis} provides a detailed analysis of how our training pipeline improves action generation and how each modality is affected under the training pipeline. Finally, \ref{sub:ablation study} reports ablation studies on decoding strategies of each modalities and robot's state in text or image's context.

\subsection{Benchmark}
\label{sub:benchmark}
\paragraph{Simulation Benchmark}

We conducted validation of diverse tasks in the simulation experiments.

\textbf{LIBERO}~\cite{libero} is a benchmark based on the Franka robotic arm, designed to evaluate lifelong learning and knowledge transfer capabilities in robots. In this work, we adopt four sub-benchmarks from LIBERO: Libero-Spatial for spatial reasoning, Libero-Object for object-centric understanding, Libero-Goal for goal-conditioned variations, and Libero-Long for long-horizon, compositional tasks. Each sub-benchmark contains 10 distinct tasks, and each task is provided with 50 tele-operated demonstrations.

\begin{figure*}[ht!]
    \centering
    \includegraphics[width=\textwidth]{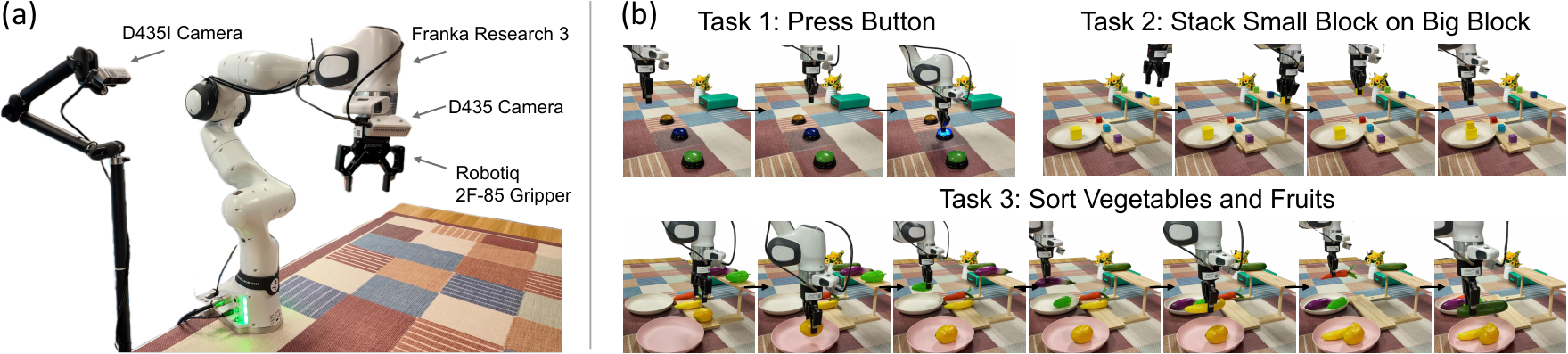}
    \caption{
        \textbf{(a) Franka real-world setup.} The Franka Research 3 robot, equipped with a D435i camera as an external camera, a D435 camera as a wrist camera, and a Robotiq 2F-85 gripper, is shown in the experimental setup.
        \textbf{(b) Task examples.} 
            \textbf{Press Button: } The robot presses the blue button.
            \textbf{Stack Small Block on Big Block:} The robot stacks the small yellow block on top of the large yellow block.
            \textbf{Sort Vegetables and Fruits:} The robot places fruits from the rack into the right plate and vegetables into the left plate.
    }
    \label{fig:franka_setting}
\end{figure*}

\textbf{RoboTwin}~\cite{robotwin, robotwin2} is a simulation benchmark designed for bimanual robotic manipulation. We employ Robotwin2.0~\cite{robotwin2} for our experiments. This simulation platform provides a multi-task data collection and evaluation framework across multiple robot embodiments.  By introducing a large variety of scenes and objects for domain randomization, it both scales up the available robot training data and enables out-of-domain, unseen evaluation. In this work, we adopt the Agilex Piper dual-arm setup and evaluate on eight representative tasks in unseen settings, which task instructions, environments and object location are unseen in training datasets. 
\paragraph{Real-world Experiments}
 We employ Franka real-world experiments for testing the in-domain action generation ability among our model and baselines. We use an Intel RealSense D435 camera mounted on the wrist and an Intel RealSense D435i camera positioned as a third-person view to provide visual observations for the model. We design three manipulation tasks—press button, stack small block on big block, and sort vegetables and fruits (see Figure~\ref{fig:franka_setting} for task illustrations). For each task and each model, we conduct 20 evaluation trials. For different tasks, we employ distinct success rate calculation methods. For press button and stack block, the task is considered successful when the entire task is completed. For sort vegetables and fruits, the success rate is calculated based on the number of objects correctly placed in the right spot.

\begin{table*}[ht!]
\centering
\begin{adjustbox}{width=0.8\linewidth} 
\begin{tabular}{lcccc}
\hline
{\textbf{Model}} & \textbf{Press Button} & \textbf{Stack Block}  & \textbf{Sort Vegetable and Fruits} & \textbf{Average} \\
\hline
$\pi_0$ \cite{black2024pi_0} & 75.0 & \textbf{70.0} & 65.0  & 70.0 \\
OpenVLA-OFT \cite{kim2025openvlaoft} & 70.0 & 50.0 & 56.0 & 58.6\\
\textbf{\methodname} & \textbf{80.0} & \textbf{70.0} & \textbf{66.0}  & \textbf{72.0} \\
\hline
\end{tabular}
\end{adjustbox}
\caption{\textbf{Main results of Franka Real-world Experiments.}}
\label{tab:franka-real}
\end{table*}
\paragraph{Baselines}
We evaluate our method against three major paradigms: (1) VLM-based VLA (e.g. OpenVLA~\cite{kim2024openvla}, OpenVLA-OFT~\cite{kim2025openvlaoft}, $\pi_0$~\cite{black2024pi_0}), which focus on semantic understanding but lack dynamic modeling; (2) Visual Prediction VLA (e.g. CoT-VLA~\cite{zhao2025cotvla}, TraceVLA~\cite{zheng2024tracevla}, DreamVLA~\cite{zhang2025dreamvla}), which emphasize future prediction but have limited task reasoning; and (3) Unified VLA based on the unified model architecture (e.g. UniVLA~\cite{wang2025univla}, WorldVLA~\cite{cen2025worldvla}), which possesses the ability to generate multiple modalities and is capable of performing visual or textual reasoning.

\subsection{Implementation Details}
\label{sub:implementation details}
\paragraph{Datasets}
For LIBERO benchmark, we use the official datasets provided from the benchmark. To verify whether our training pipeline can leverage text-based task planning to improve action generation, we focus on the long-horizon LIBERO-Long setting. We manually decompose each of the 10 tasks into a sequence of subtasks, annotate keyframes corresponding to subtask transitions for every episode, and then generate subtask planning labels for each frame using a template-based scheme.

For RoboTwin2.0 benchmark, we collect 500 expert episodes per task under the domain-randomized setting and automatically filter out trajectories exhibiting undesirable behaviors such as overly long pauses, resulting in a total of about 70k training samples. We further design an automatic subtask-annotation framework on simulation platform, based on the rule-based trajectory generation used for collecting RoboTwin expert data. We timestamp each invocation of motion, grasp, and place primitives from the skill library, and attach task-specific language descriptions for each skill function called in each task. This allows us to automatically obtain the executed action chunk, the corresponding subtask language annotation and the future image after action execution for every frame.

For each Franka real-robot task, we collect 100 demonstration trajectories via manual teleoperation.

All of our model weights are directly trained from the base weights of MMaDA~\cite{yang2025mmada} on the above datasets, differing from VLA models that are pretrained on large-scale robotic datasets and finetune on small datasets, such as OpenVLA~\cite{kim2024openvla} and $\pi_0$~\cite{black2024pi_0}. More details of our datasets can be found in Appendix~\ref{app-sec:datasets-details}.

\paragraph{Training Details}
We train our model with batch size of 128 and action chunk size of 8 across three experiments. For LIBERO benchmarks, We train separate model for each sub-benchmark with an average of 11k steps. For eight RoboTwin tasks, we adopt multi-task training and optimize action generation for about 27k steps. For Franka real-world experiments, we train for roughly 8k steps per task. For baselines in RoboTwin, we train them using the same batch size, learning rate, action chunk size and gradient steps as ours. For baselines in Franka real-world experiment, we use their default fine-tuning hyperparameters and evaluate the best checkpoint selected within 30k training steps. More details of training can be found in Appendix~\ref{app-sec:training pipeline details}.

\subsection{Evaluation Results}
\label{sub:evaluation_results}

\paragraph{Benchmark Results}

\begin{figure*}
    \centering
    \includegraphics[width=\linewidth]{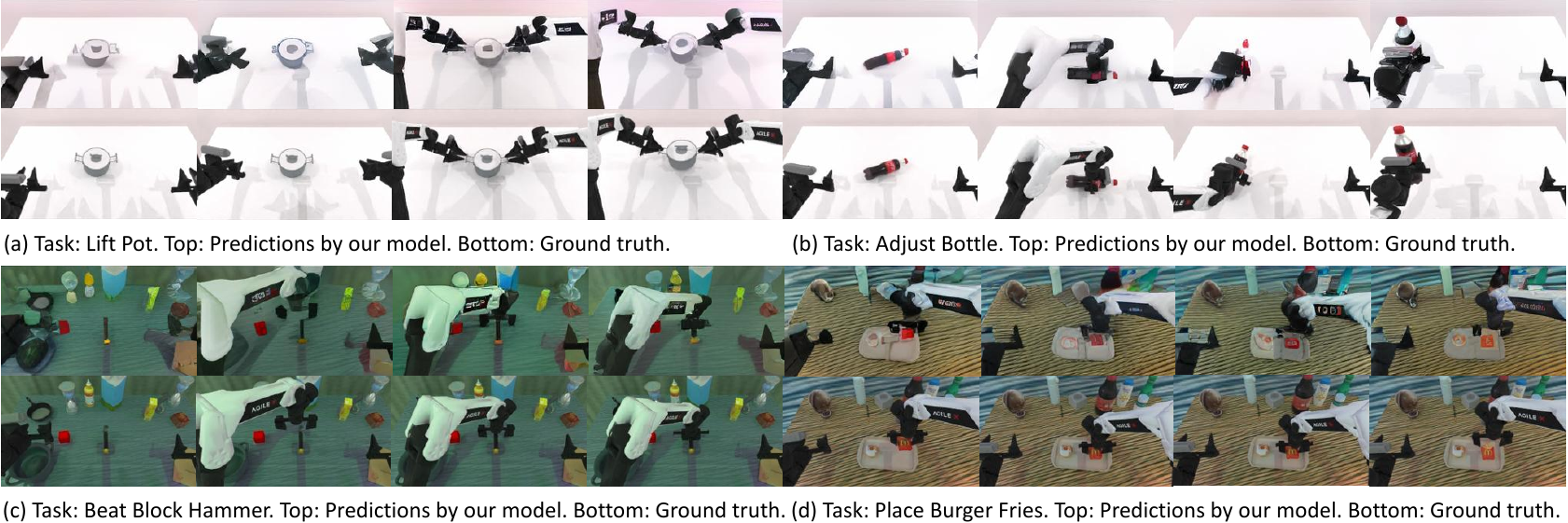}
    \caption{\textbf{Visualization of image generation by \methodname on RoboTwin unseen environments.} (a) and (b) show the generation results in clean scenes, (c) and (d) show the generation results in domain-randomized scenes. All of the scenes test the model's performance in the out-of-domain setting. The top part shows the images generated by our model, while the bottom part represents the ground truth.}
    \label{fig:robotwin_image_generation}
\end{figure*}

As shown in Table~\ref{tab:libero}, \methodname achieves the best average success rate of 96.3\%, surpassing all existing baselines. Specifically, the vanilla version of \methodname achieves success rates of 97.8\%, 99.4\%, 94.8\%, and 88.0\% on Libero-Spatial, Libero-Object, Libero-Goal, and Libero-Long respectively.
For VLM-based VLA, our approach outperforms OpenVLA by 19.8\%, $\pi_0$ + FAST by 10.8\%, $\pi_0$ by 2.1\%, and OpenVLA-OFT by 0.9\%.
For Visual Prediction VLA, our method surpasses CoT-VLA by 15.2\%, TraceVLA by 21.5\%, and DreamVLA by 3.7\%.
In the case of Unified VLA, our approach exceeds WorldVLA by 14.5\% and UniVLA by 0.8\%.
This validates the effectiveness of \methodname in terms of model architecture and training paradigms.
Furthermore, as shown in the last two rows of the table, jointly optimizing the model for both task planning and action generation during training effectively improves its capability in long-horizon planning. In particular, the success rate on Libero-Long increases from 88.0\% to 93.0\%, achieving a notable improvement of +5.0\%.

Table~\ref{tab:robotwin} presents the main results of different VLA models on the RoboTwin benchmark across eight manipulation tasks. \methodname achieved the best average performance of 52.38\%, surpassing $\pi_0$ by 4.25\% and OpenVLA-OFT by 29.25\%.

As detailed in Table~\ref{tab:franka-real}, in the real-world Franka experiments, \methodname achieved the highest average success rate of 72.0\%, far surpassing $\pi_0$ at 70.0\% and OpenVLA-OFT at 58.6\%. This demonstrates the superiority of \methodname and proves its ability to successfully tackle real-world task challenges.

\subsection{Analysis on Training Pipeline}
\label{sub:training_analysis}

\paragraph{Action Enhancement via Multimodal Learning}
We present the experimental results of multimodal learning in Table~\ref{tab:libero} and Table~\ref{tab:robotwin}. Specifically, Table~\ref{tab:libero} demonstrates the effectiveness of our training pipeline, where we simultaneously train task planning and action generation on the LIBERO-Long long-horizon tasks. Compared with the baseline trained under an identical setting but generating actions only, our approach achieves a 5.0\% improvement in success rate. Table~\ref{tab:robotwin} further compares the success rates among action-only training, text-action training, image-action training, and unified text-image-action training. Our context-shared multimodal learning approach has led to an increase in success rates for action generation, with a 3.37\% improvement when co-trained with text, a 5.62\% increase when co-trained with image, and a 9.25\% boost when all three modalities—text, image and action—were jointly trained, achieving the highest success rate of 52.38\%. This also validates the effectiveness of our training pipeline.

\paragraph{Image Quality Assessment}
Figure~\ref{fig:robotwin_image_generation} presents the generation results of \methodname(+Image) in clean and cluttered unseen scenes on RoboTwin. It can be observed that the generated images closely resemble the subgoal image, retaining key information and capable of predicting dynamic changes in the environment.

We evaluated the quality of the generated images using three metrics at Table~\ref{tab:image_quality}: PSNR~\cite{gonzalez2002digital} (Peak Signal-to-Noise Ratio), SSIM~\cite{wang2004image} (Structural Similarity Index), and LPIPS~\cite{zhang2018lpips} (Learned Perceptual Image Patch Similarity). 
Table~\ref{tab:image_quality} demonstrates that our context-shared multimodal learning pipeline enables the model to learn effective action generation during joint training while simultaneously improving the quality of future image prediction.

\begin{table}[ht!]
\centering
\caption{\textbf{Image generation quality}. We evaluated three image-generation metrics using model weights from Stage 1 and Stage 2 in \textbf{unseen} experiments for 1000 attempts. Stage 1 is trained solely on image generation objectives, while Stage 2 performs joint training on both image and action generation after Stage 1.}
\label{tab:image_quality}
\begin{tabular}{>{\raggedright\arraybackslash}p{2cm}  p{1.5cm} p{1.5cm} p{1.5cm}} 
\toprule
\textbf{\methodname} & \textbf{PSNR $\uparrow$} & \textbf{SSIM $\uparrow$} & \textbf{LPIPS $\downarrow$} \\
\midrule
Stage 1  &      12.08    &     0.79     &    0.11     \\
Stage 2 &  \textbf{14.23}    &     \textbf{0.80}      &     \textbf{0.09}    \\
\bottomrule
\end{tabular}
\end{table}

\paragraph{Text Quality Assessment}
We conduct a detailed analysis and additional experiments on the generation quality of the model's text modalities.

\begin{figure*}[!htbp]
  \centering
  \begin{minipage}[t]{0.28\textwidth}
    \centering
    \includegraphics[width=\linewidth]{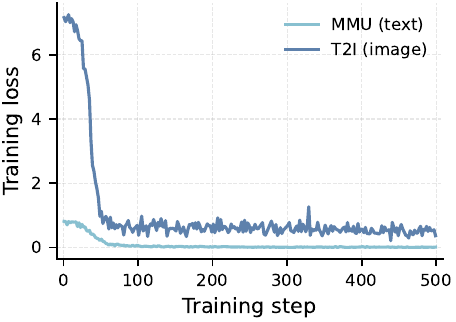}
    \vspace{2pt}
    {\small (a) Stage1}
  \end{minipage}
  \hfill
  \begin{minipage}[t]{0.67\textwidth}
    \centering
    \includegraphics[width=\linewidth]{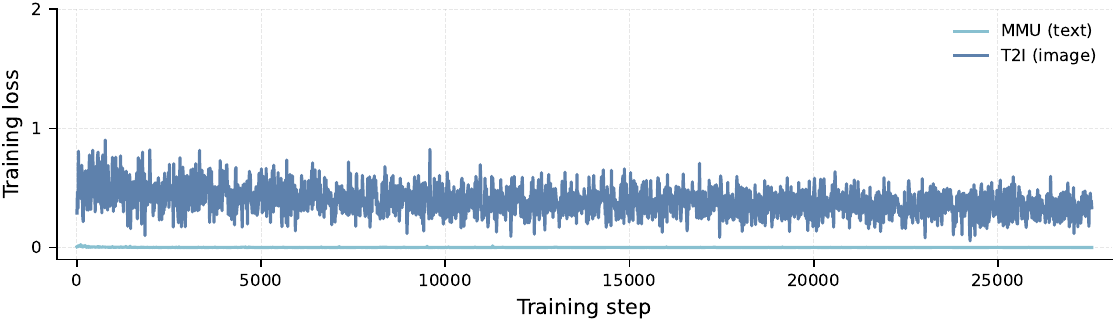}
    \vspace{2pt}
    {\small (b) Stage2}
  \end{minipage}

  \caption{\textbf{Comparison of training loss between MMU (text) and T2I (image) on two training stages.} In Stage 1, the loss of text modality rapidly converges to a value very close to $0$, while the loss of image modality quickly decreases below $1$ and then declines slowly. In Stage 2, the loss of text modality remains consistently near $0$, whereas image modality loss exhibits a slow, oscillating decline over an extended number of training steps.}
  \label{fig:loss_mmu_t2i_two}
\end{figure*}
For text modality, we compare the task planning accuracy of MM-ACT (+Text) at Stage 1 and Stage 2 on clustered unseen scenes. The evaluation dataset is similarly collected through our RoboTwin data pipeline, with the key distinction that we select unseen scenes and object spatial arrangements from the training dataset. To evaluate the correctness between our model's output and the ground truth, we leverage GPT-4o~\cite{hurst2024gpt} with the prompt shown in Figure~\ref{fig:prompt_llm_judge_simple}.
\begin{figure}[!htbp]
\centering
\begin{minipage}{0.95\linewidth}
\ttfamily\small
You are a judge for embodied task planning. Your job:

- Compare an agent's plan (agent\_plan) with a reference plan (ground\_truth).

- Decide whether they are consistent in terms of task planning and decisions for the current task.

- The two plans do NOT need to be exactly the same; similar intent and decision logic are enough.

Requirements:

- If they are consistent, answer exactly ``yes''.

- If they are not consistent, answer exactly ``no''.

- Do not output anything else.

Here is the data: 

agent\_plan: \{agent\_plan\}. 

ground\_truth: \{ground\_truth\}
\rmfamily
\end{minipage}
\caption{\textbf{Prompt used for LLM judge in our experiments.} $\texttt{\{agent\_plan\}}$ is our model's output, \texttt{\{ground\_truth\}} denotes the task planning annotation corresponding to the evaluation sample.}
\label{fig:prompt_llm_judge_simple}
\end{figure}

We conduct evaluations for 1,000 attempts, and the results are presented in Table~\ref{tab:text_quality}. In Figure~\ref{fig:visualization-of-text-generation}, we present several comparisons between the task planning outputs generated by our model and the corresponding ground truth annotations.
\begin{table}[!htbp]
\centering
\caption{\textbf{Text generation quality.} Stage 1 is trained solely on text-generation objectives, while Stage 2 performs joint training on both text and action generation after Stage 1. Accuracy is defined as the proportion of evaluations in which LLM judge outputs "yes" among all evaluations.}
\label{tab:text_quality}
\begin{tabular}{>{\raggedright\arraybackslash}p{2cm}  p{2cm}} 
\toprule
\textbf{\methodname} & \textbf{Acc (\%)} \\
\midrule
Stage 1  & \textbf{81.5}     \\
Stage 2 &  68.7    \\
\bottomrule
\end{tabular}
\end{table}

This indicates that our model acquires strong task planning capabilities during the initial Stage 1, where the text modality is trained independently. However, after training with action modality in Stage 2, the text generation performance deteriorates, which is inconsistent with the results observed in the image generation modality. We visualize the training curves of the text and image modalities during Stage 1 and Stage 2 in Figure~\ref{fig:loss_mmu_t2i_two}.

As observed, the training loss for text modality during Stage 1 rapidly approaches $0$ in approximately 100 steps, indicating a near-perfect fitting in our dataset. In contrast, the loss for image modality continues to decrease consistently throughout both Stage 1 and Stage 2 training. This suggests that text modality is prone to overfitting with increased training steps, resulting in decreased generalization performance on unseen scenarios. Meanwhile, the slower fitting process of the image modality allows it to continuously benefit from our training pipeline, achieving steady improvements.

\subsection{Ablation Study}

\label{sub:ablation study}

\paragraph{Action Decoding Strategy}
\label{parag:action decoding strategy}
Different from the re-mask decoding strategy employed for image and text modalities, we adopt a one-step parallel decoding strategy for action prediction, significantly accelerating the speed and reducing the frequency of model forward processes. To comprehensively analyze the effectiveness and computational efficiency of these two decoding strategies, we conduct comparative experiments under two settings: action chunk sizes of 8 and 16. The experimental results are summarized as Table~\ref{tab:ablation action decoding strategy}:
\begin{table}[ht!]
\centering
\caption{\textbf{Action decoding strategy}. ``one-step PD" denotes that the model generates actions using a one-step parallel decoding strategy. ``Re-mask PD" indicates that the model performs multiple forward processes with iterative re-masking. ``cs" refers to the chunk size, and ``t" denotes the number of forward steps.}
\label{tab:ablation action decoding strategy}
\begin{tabular}{>{\raggedright\arraybackslash}p{3.5cm}  p{2.5cm} p{1cm}} 
\toprule
\textbf{\methodname} & \textbf{Overall Avg (\%)} & \textbf{Time}\\
\midrule
one step PD, cs=8, t=1  &      \textbf{43.13}    &      0.22s         \\
re-mask PD, cs=8, t=6 &  42.38\textcolor{red}{(-0.75\%)}    & 1.06s  \\
one step PD, cs=16, t=1  &      43.75    &          0.23s     \\
re-mask PD, cs=16, t=6 &  \textbf{56.75}\textcolor{blue}{(+13.00\%)}    &  1.06s \\
\bottomrule
\end{tabular}
\end{table}
As illustrated in table, when action chunk size is set to 8, the re-mask decoding strategy does not enhance action generation performance but instead leads to nearly a five-fold increase in inference time. However, when the action chunk size is increased to 16, incorporating the re-mask strategy indeed improves action generation success rate, albeit still accompanied by a significant rise in inference duration. This indicates that the re-mask parallel decoding strategy yields more pronounced improvements in generation quality when applied to tasks requiring longer parallel token sequences.

Considering the real-time control requirements in robotic tasks, we ultimately select the one-step parallel decoding strategy with an action chunk size of 8 for action generation, enabling a high generation frequency of up to 40 Hz (5Hz per action chunk).

\paragraph{Text and Image Decoding Strategies}
The ablation study on decoding strategies of text and image modalities is presented in Table~\ref{tab:ablataion_decoding_strategy}, demonstrates that during action generation training, both one-step parallel decoding and re-mask decoding strategies for text and image modalities can provide beneficial improvements to action generation performance. Besides, employing the re-mask decoding strategy in the image modality provides greater benefits compared to the one-step decoding strategy, potentially because the re-mask decoding strategy aligns more closely with the pre-training paradigm of the base model.
\begin{table}[!htbp]
  \centering
  \small
  \caption{\textbf{Ablation on decoding strategy for text and image modalities in RoboTwin2.0 Tasks.}}
  \label{tab:ablataion_decoding_strategy}
  \begin{tabular}{lccc}
    \hline
    \textbf{Model} & \textbf{Decoding Strategy} & \textbf{SR (\%)} \\
    \hline
    MM-ACT (+Text)  & re-mask        & 46.50(\textcolor{blue}{+3.37\%}) \\
    MM-ACT (+Text)  & one-step        & \textbf{46.63}(\textcolor{blue}{+3.50\%}) \\
    MM-ACT (+Image)   & re-mask       & \textbf{48.75}(\textcolor{blue}{+5.62\%}) \\
    MM-ACT (+Image)   & one-step       & 46.13(\textcolor{blue}{+3.00\%}) \\
    MM-ACT (Vanilla) & - & 43.13 \\
    \hline
  \end{tabular}
\end{table}

\paragraph{State in Text or Image's Context}
The ablation study on the inclusion of robot's state in context is presented in Table~\ref{tab:state_ablation}, reveals that whether incorporating robot's state in the context of text and image modalities could both provide improvements to action generation performance, though enhancement varies across the two modalities. In text modality, including the robot's state as part of the context reduces the beneficial effect on action generation. In contrast, in image modality, incorporating the robot's state into the context enhances the beneficial effect on action generation, possibly because image generation aligns more finely with action generation, allowing closer context to mutually reinforce performance improvements.

\begin{table}[!htbp]
  \centering
  \small
  \caption{\textbf{Ablation study on the inclusion of robot's state in text's or image's context in RoboTwin2.0 Tasks.}}
  \label{tab:state_ablation}
  \begin{tabular}{lccc}
    \hline
    \textbf{Model} & \textbf{State} & \textbf{SR (\%)} \\
    \hline
    MM-ACT (+Text)  & without        & \textbf{46.50}(\textcolor{blue}{+3.37\%}) \\
    MM-ACT (+Text)  & with        & 43.50(\textcolor{blue}{+0.37\%}) \\
    MM-ACT (+Image)   & without       & 48.75(\textcolor{blue}{+5.62\%}) \\
    MM-ACT (+Image)   & with       & \textbf{51.50}(\textcolor{blue}{+8.37\%}) \\
    MM-ACT (Vanilla) & - & 43.13 \\
    \hline
  \end{tabular}
\end{table}

\section{Conclusion}

We present MM-ACT, a unified Vision-Language-Action model that generates text, image, and robot's action through a shared discrete token space and parallel decoding with bidirectional attention. This unified architecture eliminates hybrid decoding complexity and enables simple training pipeline designs. To enhance cross-modal learning, we propose Context-Shared Multimodal Learning, which jointly supervises all modalities from the same context, fostering synergy among task planning, future image prediction and action generation. MM-ACT achieves strong results across benchmarks: 96.3\% on LIBERO, 52.38\% on RoboTwin2.0 eight tasks, and 72.0\% on Franka real-world tasks, with context-shared multimodal learning pipeline yielding +9.25\% gains in out-of-domain performance. Ablation studies validate the trade-offs between one-step parallel decoding and re-mask strategy, and further reveal that our training pipeline improves both action and image generation. Overall, MM-ACT offers a compact and effective framework for discrete multimodal generation in embodied agents, paving the way for more modalities in future scaling.

\clearpage
{
    \small
    \bibliographystyle{ieeenat_fullname}
    \bibliography{main}
}

\clearpage
\setcounter{page}{1}
\title{\Large\bf Appendix}
\maketitle
\appendix
\section{Dataset Construction \& Annotation Details}
\label{app-sec:datasets-details}
In both simulated and real-world experiments with Franka arm, we adopt the robot end-effector's delta pose as the action representation. For RoboTwin experiments, we directly use the robot end-effector's absolute pose as the action.

For text annotations in LIBERO Long dataset, we first manually compose several sub-task categories corresponding to each of the 10 tasks. Then, we manually match the sub-task annotation within its corresponding episode and annotate the specific frames indicating transitions between sub-tasks. Thus, each episode is annotated into several sections by these specific frames, with each section corresponding to specific descriptions.

\begin{figure}[!htbp]
  \centering
  \small
  \setlength{\tabcolsep}{3pt}
  
  \begin{tabular}{@{}llp{0.6\linewidth}@{}}
    \hline
    \textbf{ID} & \textbf{Item} & \textbf{Example Content} \\
    \hline
    \textbf{[1]} & \texttt{instructions} 
    & Grab the black and yellow hammer grip, then hit the block. \\
    \textbf{[2]} & \texttt{planning\_text} 
    & Grasp the hammer with the right arm $\rightarrow$ Lift the hammer upwards $\rightarrow$ Move the hammer over the block $\rightarrow$ place it down to beat the block. \\
    \textbf{[3]} & \texttt{history\_text} 
    & Grasp the hammer with the right arm, Lift the hammer upwards \\
    \textbf{[4]} & \texttt{subtask\_text} 
    & Move the hammer over the block, place it down to beat the block. \\
    \hline
  \end{tabular}

  \vspace{0.4em}
  \includegraphics[width=\linewidth]{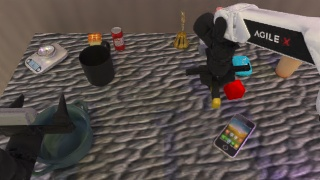}

  \fbox{%
    \parbox{\linewidth}{%
      \textbf{Resulting Description:}\\[2pt]
         My task is \textbf{[1]} Grab the black and yellow hammer grip, then hit the block. I need to finish this task by \textbf{[2]} Grasp the hammer with the right arm, Lift the hammer upwards, Move the hammer over the block and place it down to beat the block. Currently, I have finished \textbf{[3]} Grasp the hammer with the right arm, Lift the hammer upwards. So now I should continue to \textbf{[4]} Move the hammer over the block and place it down to beat the block.
    }%
  }
\caption{\raggedright
\textbf{Task planning annotation example in RoboTwin2.0.}
\texttt{instructions}, \texttt{planning\_text},
\texttt{history\_text}, and \texttt{subtask\_text}
are concatenated into a single annotation.
}
  \label{fig:annotation-template}
\end{figure}

The expert data from RoboTwin is annotated through a low-level, rule-based path planning procedure, which involves a fixed sequence of skill function call for each task. Building on this predefined function call sequence, we label task-relevant sub-tasks for each function call and timestamped the completion of each call during expert data generation. This approach enables us to automatically obtain expert data with corresponding task-level language annotations directly from the automated data collection pipeline. Figure~\ref{fig:annotation-template} illustrates an example of how we construct the task planning annotation.
Ultimately, we expand these annotations into structured task planning texts according to predefined templates, which were then utilized for the final training. This method ensures that each frame in our expert dataset includes textual task-planning annotations.

For future image prediction annotations, we directly select the frame after executing the corresponding action chunk as the ground truth for future image prediction. This method allows us to leverage the temporal nature of the original dataset without requiring additional manual annotations. Based on the aforementioned approach, we can utilize an automated data collection pipeline to gather corresponding frame-by-frame action, text, and image data from Robotwin2.0 simulation for training our model.

\section{More details of Re-mask Parallel Decoding}
\label{app-sec:details of remask}
In our framework, the re-mask strategies for text and image modalities differ in design. For text modality, the mask schedule function $f_{\mathrm{modal}}$ is simply set linear as: $f_{\mathrm{modal}}(t)=t$. For image modality, we adopt cosine schedule function, set $f_{\mathrm{modal}}=\cos\!\left( \frac{\pi}{2}\, (1-t) \right)$.

During inference, the number of masked tokens to predict at each timestep for both text and image modalities is determined according to the noise schedule function. For text and image modalities, predicted tokens with higher confidence are preferentially selected for retain, while predicted tokens with lower confidence are re-masked. In our practice, we set the temperature to 0 and do not incorporate classifier-free guidance (CFG). 

For action modality, we introduce a one-step parallel decoding strategy during both training and inference, requiring the model to predict all masked tokens within a single forward process. Regarding the re-mask decoding strategy, which we also apply to the action modality, our implementation aligns consistently with that of image modality.

\section{Training pipeline details}
\label{app-sec:training pipeline details}
For LIBERO benchmark, all training is conducted with a batch size of 128 and a learning rate of $5\times10^{-5}$. Among the reported model weights, LIBERO Object is trained for about 9k steps, LIBERO Spatial for about 7.5k steps, LIBERO Goal for about 8.5k steps, and LIBERO Long for about 17.5k steps. For "+Text" in LIBERO Long, we first train the text generation for one epoch (about 800 steps), with $\lambda_\mathrm{mmu}=1$ in this stage. Subsequently, we jointly train the text and action generation modalities, assigning a weight of $0.05$ to $\lambda_\mathrm{mmu}$ and a weight of $1$ to $\lambda_\mathrm{mm2a}$.

For RoboTwin benchmark, we maintain the same batch size and learning rate as described above. For Vanilla model, training begins from the base model weights and continues for approximately 27k steps (5 epochs). For both the "+Text" and "+Image" models, we initially train each modality independently for 500 steps, assigning $\lambda_\mathrm{mmu}$ or $\lambda_\mathrm{t2i}$ to $1$ during this stage. We then jointly train the modality together with actions for approximately 27k steps (5 epochs), assigning a modality weight of $0.1$ for either $\lambda_\mathrm{mmu}$ or $\lambda_\mathrm{t2i}$, and with $\lambda_\mathrm{mm2a}=1$. For the "+Text\&Image" model, we first jointly train the text and image modalities for 500 steps, setting the weights of both $\lambda_\mathrm{mmu}$ and $\lambda_\mathrm{t2i}$ to $1$. Subsequently, we train all three modalities together, assigning a weight of $1$ to $\lambda_\mathrm{mm2a}$, $0.05$ to both $\lambda_\mathrm{mmu}$ and $\lambda_\mathrm{t2i}$.

\section{Robotic Embodiments in Simulation \& Real-World}
\label{app-sec:embodiments}
\begin{figure}[htbp]
    \centering
    \includegraphics[width=\linewidth]{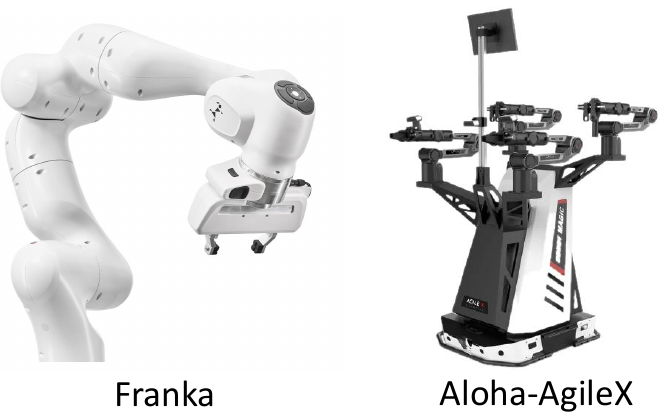}
    \caption{\textbf{Embodiments used in simulation and real-world experiments.}}
    \label{fig:embodiments}
\end{figure}

In the LIBERO and real-world experiments, we use Franka as the embodiment; in the RoboTwin experiment, we use Aloha-AgileX as the embodiment. The specific embodiments are visualized in Figure~\ref{fig:embodiments}. 
Franka Research 3 is a force-sensitive robotic system designed for robotics and artificial intelligence research. The system features a 7 degree-of-freedom (DoF) arm with integrated torque and force sensors at each joint, supporting control frequencies of up to 1 kHz. 
Aloha‑AgileX is a robotic platform integrating a mobile base with dual manipulator arms, enabling whole‑body teleoperation of both the base and the arms. The system supports synchronous coordination of the differential‑drive chassis and the bimanual arms, thus expanding the operational workspace beyond static manipulators. By combining the base’s linear and angular velocity control with the manipulators’ multiple joint actuations (e.g., the original ALOHA system features approximately 14DoF), the platform realizes a high‑dimensional action space for research in mobile manipulation, bimanual coordination, and simulation‑to‑real‑world transfer. 

\section{Task Visualizations on RoboTwin2.0}
\label{app-sec:visualization-task}

\begin{figure*}
    \centering
    \includegraphics[width=\textwidth]{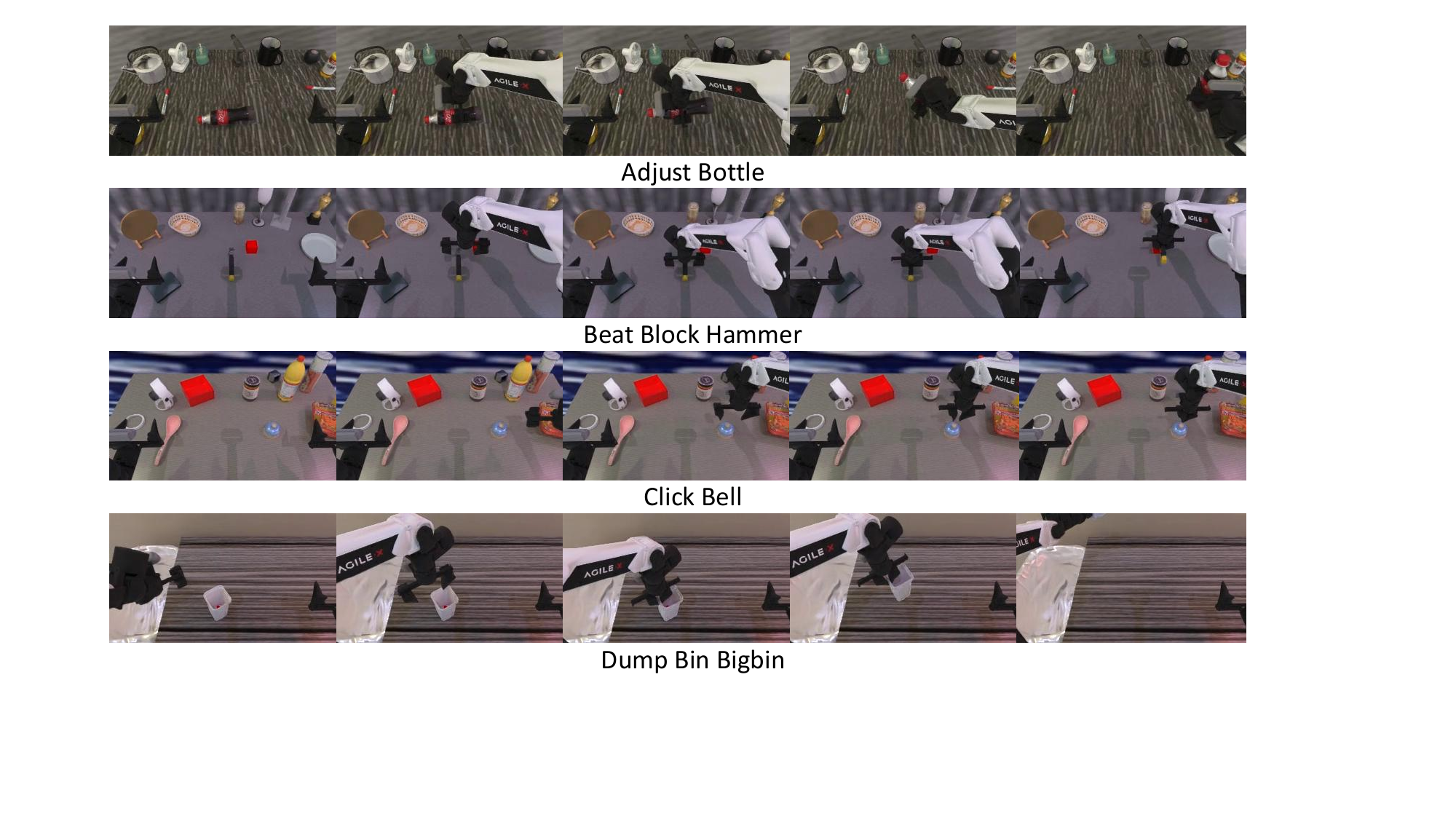}\\[0pt]
    \includegraphics[width=\textwidth]{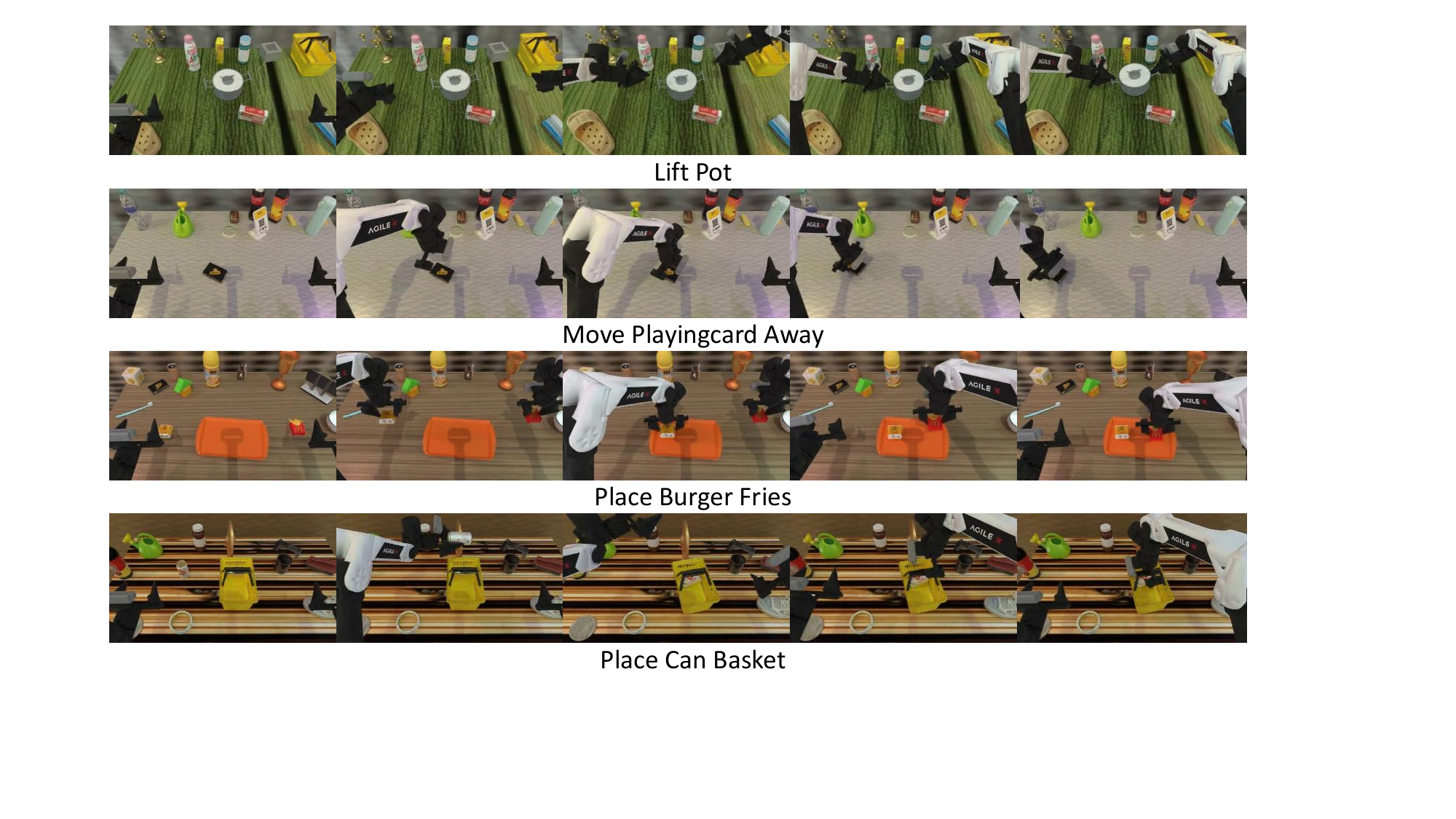}
    \caption{\textbf{Visualization of eight tasks in RoboTwin2.0.}}
    \label{fig:robotwintasks_examples}
\end{figure*}

In our experiments, we used eight tasks from the RoboTwin 2.0 simulation benchmark. The following are detailed descriptions of each task, along with visualizations with domain randomization, as shown in Figure~\ref{fig:robotwintasks_examples}:

\begin{itemize}
    \item \textbf{Adjust Bottle}: Pick up the bottle on the table and place it upright using the correct arm.
    \item \textbf{Beat Block Hammer}: There is a hammer and a block on the table; use the arm to grab the hammer and strike the block.
    \item \textbf{Click Bell}: Click the top center of the bell on the table.
    \item \textbf{Dump Bin Bigbin}: Grab the small bin and pour the balls into the big bin.
    \item \textbf{Lift Pot}: Use the arm to lift the pot.
    \item \textbf{Move Playingcard Away}: Use the arm to pick up the playing card and move it away from the table. For example, if the playing card is on the outward side of the table, you should move it further outward.
    \item \textbf{Place Burger Fries}: Use both arms to pick up the burger and fries and place them onto the tray.
    \item \textbf{Place Can Basket}: Use one arm to pick up the can and place it into the basket, while the other arm lifts the basket.
\end{itemize}

\begin{figure*}[htbp]
    \centering
    \begin{minipage}[t]{0.45\textwidth}
        \centering
        \includegraphics[width=\textwidth]{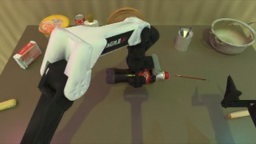}
        \caption*{
            \textbf{Task: }Adjust Bottle\\
            \textbf{Generated Text: }My task is Lift the Coca-Cola bottle from the table upright. I need to finish this task by Grasp the bottle with the left arm, Move the bottle to the target position. So now I should continue to Grasp the bottle with the left arm.\\
            \textbf{Ground Truth: }My task is Lift the Coca-Cola bottle from the table upright. I need to finish this task by Grasp the bottle with the left arm, Move the bottle to the target position. So now I should continue to Grasp the bottle with the left arm.
        }

    \end{minipage}%
    \hspace{0.05\textwidth}  %
    \begin{minipage}[t]{0.45\textwidth}
        \centering
        \includegraphics[width=\textwidth]{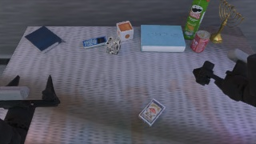}
        \caption*{
            \textbf{Task: }Move Playingcard Away\\
            \textbf{Generated Text: }My task is Place the blue carton for playingcards away from the table outward.. I need to finish this task by Grasp the with the right arm, Move the playing cards away the right arm. So now I should continue to Grasp the playing with the right arm.\\
            \textbf{Ground Truth: }My task is Place the blue carton for playingcards away from the table outward.. I need to finish this task by Grasp the playing cards with right arm, Move the playing cards away with right arm. So now I should continue to Grasp the playing cards with right arm.
        }
    \end{minipage}
    
    \vspace{0.2cm} 
    \rule{\textwidth}{0.4pt}  %
    \vspace{0.2cm}  
    
    \begin{minipage}[t]{0.45\textwidth}
        \centering
        \includegraphics[width=\textwidth]{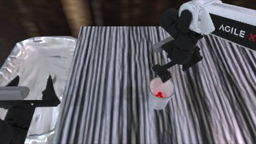}
        \caption*{
            \textbf{Task: }Dump Bin Bigbin\\
            \textbf{Generated Text: }My task is Take the compact tabletop trashbin and pour the balls out.. I need to finish this task by Grasp the trash bin with the right arm, Move the bin to the middle of the middle table and place it down, Grasp the trash bin with the left arm, and move the right arm back to the origin. Currently, and, Move the bin to the middle of the middle table and place it down. So now I should continue to Grasp the trash bin with the right arm.\\
            \textbf{Ground Truth: }My task is Take the compact tabletop trashbin and pour the balls out.. I need to finish this task by Grasp the trash bin with the right arm, Move the trash bin to the middle of the table and place it down, Grasp the trash bin with the left arm, and move the right arm back to the origin, Move the trash bin over the big dustbin, and shake it to dump the garbage inside. So now I should continue to Grasp the trash bin with the right arm.
        }
    \end{minipage}%
    \hspace{0.05\textwidth}
    \begin{minipage}[t]{0.45\textwidth}
        \centering
        \includegraphics[width=\textwidth]{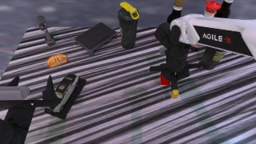} %
        \caption*{
            \textbf{Task: }Beat Block Hammer\\
            \textbf{Generated Text: }My task is Use the right arm to grab the hammer with claw-shaped end and beat block. I need to finish this task by Grasp the hammer with the right arm, Move the hammer to the target position, Lift the hammer with the arms upward. Currently, I have finished Grasp the hammer with the right arm, Move the hammer to the target position. So I should continue to Lift the hammer with the arms down.\\
            \textbf{Ground Truth: }My task is Use the right arm to grab the hammer with claw-shaped end and beat block. I need to finish this task by Grasp the hammer with the right arm, Lift the hammer upwards, Move the hammer over the block and place it down to beat the block. Currently, I have finished Grasp the hammer with the right arm, Lift the hammer upwards. So now I should continue to Move the hammer over the block and place it down to beat the block.
        }
    \end{minipage}

    \vspace{0.2cm}
    \rule{\textwidth}{0.4pt}  %
    \caption{\textbf{Visualization of text generation by MM-ACT on RoboTwin environments.} In each example, the first part is the task name, the second part is the text generated by our model, and the third part is the ground truth.}
    \label{fig:visualization-of-text-generation}
\end{figure*}

\end{document}